\documentclass[conference]{IEEEtran}
\IEEEoverridecommandlockouts
\usepackage{cite}
\usepackage{amsmath,amssymb,amsfonts}
\usepackage{algorithmic}
\usepackage{graphicx}
\usepackage{textcomp}
\usepackage{xcolor}
\usepackage{hyperref}
\usepackage{tabularx} 
\usepackage{arydshln}
\usepackage{multirow}
\usepackage{booktabs} 
\def\BibTeX{{\rm B\kern-.05em{\sc i\kern-.025em b}\kern-.08em
    T\kern-.1667em\lower.7ex\hbox{E}\kern-.125emX}}
\begin{document}

\title{PhysMamba: Synergistic State Space Duality Model for Remote Physiological Measurement}
\author{
    %Authors
    % All authors must be in the same font size and format.
    % Written by AAAI Press Staff\textsuperscript{\rm 1}\thanks{With help from the AAAI Publications Committee.}\\
    % AAAI Style Contributions by Pater Patel Schneider,
    % Sunil Issar,\\
    % J. Scott Penberthy,
    % George Ferguson,
    % Hans Guesgen,
    % Francisco Cruz\equalcontrib,
    % Marc Pujol-Gonzalez\equalcontrib
    \IEEEauthorblockN{
    Zhixin Yan\textsuperscript{\rm 1},
    Yan Zhong\textsuperscript{\rm 1},
    Hongbin Xu\textsuperscript{\rm 1},
    Wenjun Zhang\textsuperscript{\rm 1},
    Shangru Yi\textsuperscript{\rm 1},
    Lin Shu\textsuperscript{\rm 1},
    Wenxiong Kang\textsuperscript{\rm 1}
    }
    \IEEEauthorblockA{\IEEEauthorrefmark{1}South China University of Technology, Guangzhou, China}
}
 % \thanks{South China University of Technology, Guangzhou, China}
% \affiliations{
%     %Afiliations
%     \textsuperscript{\rm 1} South China University of Technology, Guangzhou, China\\
% }
% \IEEEauthorblockA{\textit{South China University of Technology, Guangzhou, China}

% \author{Anonymous ICME submission}

\maketitle

\begin{abstract}
Remote Photoplethysmography (rPPG) enables non-contact physiological signal extraction from facial videos, offering applications in psychological state analysis, medical assistance, and anti-face spoofing. However, challenges such as motion artifacts, lighting variations, and noise limit its real-world applicability. To address these issues, we propose PhysMamba, a novel dual-pathway time-frequency interaction model based on Synergistic State Space Duality (SSSD), which for the first time integrates state space models with attention mechanisms in a dual-branch framework. Combined with a Multi-Scale Query (MQ) mechanism, PhysMamba achieves efficient information exchange and enhanced feature representation, ensuring robustness under noisy and dynamic conditions. Experiments on PURE, UBFC-rPPG, and MMPD datasets demonstrate that PhysMamba outperforms state-of-the-art methods, offering superior accuracy and generalization. This work lays a strong foundation for practical applications in non-contact health monitoring, including real-time remote patient care. The code is available at \href{https://anonymous.4open.science/r/PhysMamba-E714/}{https://anonymous.4open.science/r/PhysMamba-E714/}.
\end{abstract}

\begin{IEEEkeywords}
Remote Photoplethysmography, Heart Rate Estimation, Mamba, State Space Duality
\end{IEEEkeywords}

\section{Introduction} \label{sec:intro}

Photoplethysmography (PPG) is a widely-used technique for estimating vital physiological indicators such as heart rate, heart rate variability, and respiratory rate, which are crucial for applications like sleep monitoring, medical assistance, and fatigue assessment. While traditional PPG relies on contact-based sensors like ECG monitors, its limited scalability and user inconvenience have driven the development of non-contact alternatives. Remote photoplethysmography (rPPG) leverages cameras to estimate physiological signals by capturing subtle blood flow changes in the skin \cite{choi2024fusion}, unlocking potential applications in remote health monitoring, wearable devices, and psychological state analysis.

Despite its promise, deploying rPPG in real-world environments faces significant challenges, including motion artifacts, lighting variations, and environmental noise. These factors disrupt signal consistency and hinder the generalization of rPPG systems, limiting their practical adoption in scenarios such as telemedicine and stress monitoring.

Traditional rPPG methods, including signal processing techniques like Independent Component Analysis (ICA) and chrominance-based projections \cite{li2014remote, tulyakov2016self}, perform well in controlled environments but struggle under dynamic conditions due to their reliance on handcrafted features. Recent deep learning approaches, such as CNN-based models \cite{chen2018deepphys}, have improved motion artifact handling, while Transformer-based architectures like PhysFormer \cite{yu2022physformer} enhance temporal dependency modeling. However, these models suffer from high computational complexity, making them less practical for real-time applications. State Space Models (SSMs) have emerged as a computationally efficient alternative, as demonstrated by RhythmMamba \cite{zou2024rhythmmamba}, which captures quasi-periodic patterns in physiological signals with low computational cost. Yet, existing SSM-based methods face limitations in generalization and parallel processing, particularly in cross-dataset scenarios.

\begin{figure}[t]
\centering
\includegraphics[scale=0.15]{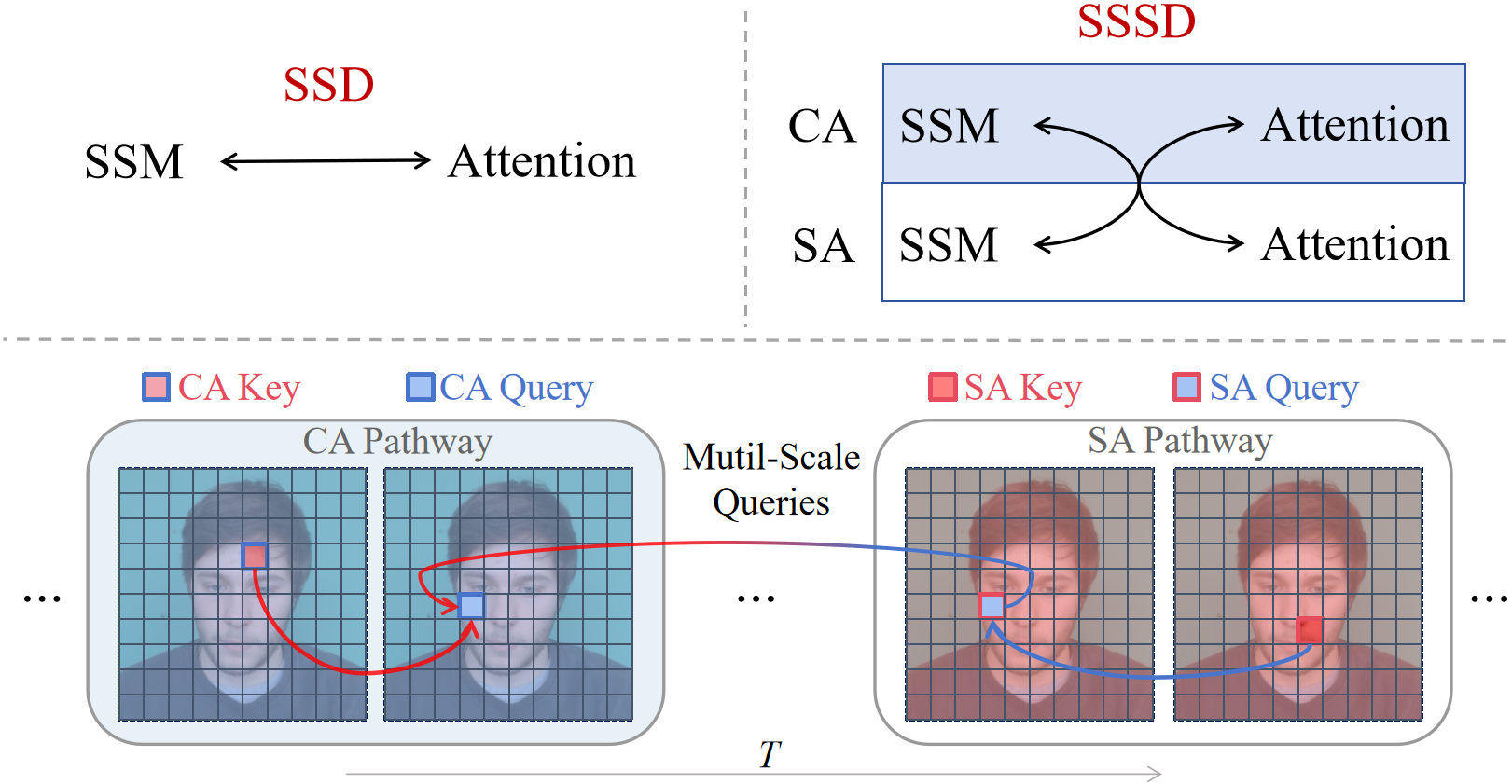}
\caption{Comparison between SSD and SSSD. The traditional SSD combines SSM and attention mechanisms, while our enhanced SSSD introduces Multi-Scale Queries for efficient information interaction between the two pathways.}
\label{graph abstract}
\end{figure}

Building on these advancements, we propose \textbf{PhysMamba}, a novel dual-pathway time-frequency interaction model that integrates \textbf{Synergistic State Space Duality (SSSD)} for rPPG estimation. This is the first framework to combine the efficiency of state space models with the feature extraction capabilities of attention mechanisms in a dual-pathway architecture. By introducing a Multi-Scale Query (MQ) mechanism, PhysMamba achieves efficient information sharing between pathways, enabling robust feature representation under noisy and dynamic conditions.

Comprehensive experiments on the PURE, UBFC-rPPG, and MMPD datasets demonstrate that PhysMamba achieves state-of-the-art performance in both intra-dataset and cross-dataset evaluations. These results highlight its superior robustness, accuracy, and generalization, paving the way for practical applications in non-contact health monitoring systems, including wearable devices and real-time telemedicine.

The main contributions of this work are summarized as follows: \begin{itemize}\item We propose PhysMamba, an efficient rPPG estimation method via State Space Duality in a dual-pathway time-frequency interactive network. This is the first attempt handling rPPG estimation via SSD, creating technological breakthroughs.
\item To enhance information exchange in dual-pathway model, we develop a simple state space fusion module using Mutil-Scale Queries for effective lateral information sharing. \item Extensive experiments on three public rPPG datasets (PURE, UBFC-rPPG, and MMPD) validate the superior performance of PhysMamba, achieving state-of-the-art results in both controlled and complex real-world scenarios. \end{itemize}

\begin{figure*}[t] 
\centering 
\includegraphics[scale=0.33]{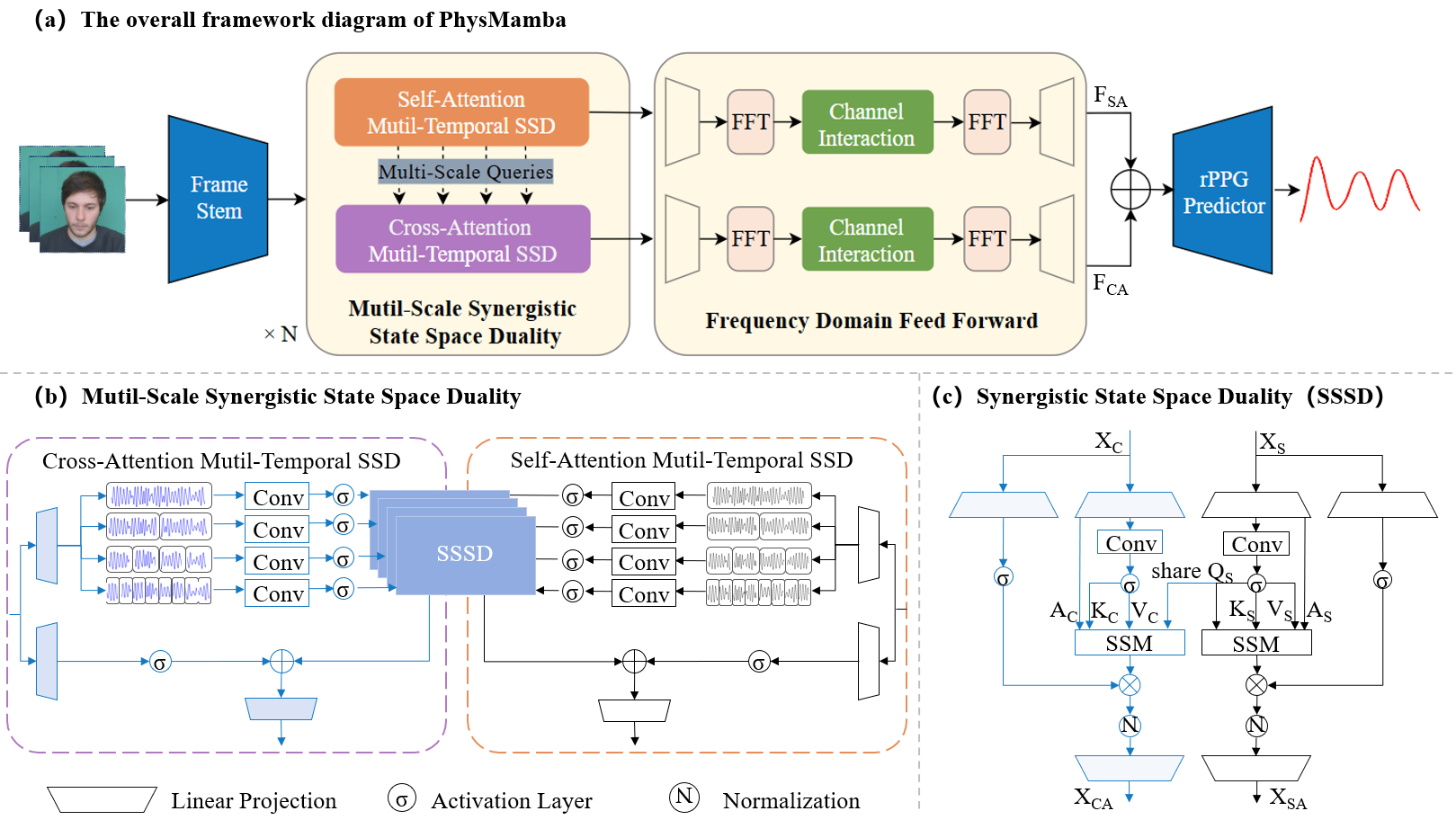} 
\caption{Overall framework diagram of the PhysMamba model. The model comprises the Frame Stem, Multi-Scale Synergistic State Space Duality (SSSD), Frequency Domain Feed-Forward (FDF), and rPPG Predictor, integrated into two pathways: the Self-Attention Pathway and the Cross-Attention Pathway.} 
\label{PhysMamba frame} 
\end{figure*}

\section{Related Work}

\subsection{rPPG Estimation}

Traditional rPPG methods, such as Independent Component Analysis (ICA) and chrominance subspace projection \cite{li2014remote, wang2016algorithmic}, perform well under controlled conditions but degrade significantly in dynamic environments due to sensitivity to motion artifacts and lighting variations \cite{niu2018synrhythm}. These approaches also rely heavily on handcrafted features, limiting their adaptability in real-world scenarios.
Deep learning has advanced rPPG estimation by enabling end-to-end frameworks. Models like DeepPhys \cite{chen2018deepphys} improved motion artifact handling through CNN architectures but struggled to capture long-term dependencies. Transformer-based methods, such as PhysFormer \cite{yu2022physformer}, leveraged self-attention mechanisms for global dependency modeling but suffered from high computational complexity, making them impractical for real-time applications.

\subsection{State Space Model}
To enhance contextual information capture in long sequences, SSMs have gained significant interest. The Mamba SSM\cite{gu2023mamba} maintains linear time complexity while effectively capturing long-term dependencies, and has been applied in fields such as time series analysis\cite{ma2024fmamba}, medical image segmentation\cite{ma2024u}, and video understanding\cite{li2024videomamba}. Recently, Mamba has been used for rPPG estimation, achieving good results with low computational complexity. RhythmMamba\cite{zou2024rhythmmamba}, an SSM-based method, robustly identifies quasi-periodic patterns in rPPG signals using a multi-phase learning framework and a frequency-domain feedforward mechanism. However, Mamba-1 had limitations in efficiency and generalization due to its dependency on previous states. Mamba-2\cite{dao2024transformers} introduced the State Space Duality framework, enhancing SSMs' linear time complexity with attention mechanisms' feature representation capabilities, improving efficiency and generalization for complex physiological signal estimation tasks like rPPG.

\section{Methodology}

\subsection{Overall Framework}

PhysMamba is an end-to-end model designed to process raw facial video data and estimate rPPG signals directly. The framework, illustrated in Figure \ref{PhysMamba frame}, consists of two pathways: the Self-Attention (SA) Pathway and the Cross-Attention (CA) Pathway. These pathways incorporate three core modules: the Frame Stem for temporal feature extraction, the Multi-Scale Synergistic State Space Duality (SSSD) module for robust temporal representation, and the Frequency Domain Feed-Forward (FDF) network for frequency enhancement. Outputs from these pathways are fused in the rPPG Predictor to generate final signal estimations. The dual-pathway architecture can be likened to a collaborative system where one pathway focuses on internal temporal consistency (SA) and the other facilitates cross-feature interactions (CA), ensuring robustness in diverse conditions.

\subsection{Frame Stem}

The Frame Stem extracts temporal features while reducing spatial noise from input video data $X \in \mathbb{R}^{3 \times T \times H \times W}$. By computing frame differences, subtle temporal variations are highlighted, which are particularly relevant for physiological signal estimation. The processed features are represented as: \begin{equation} X_{\text{Stem}} = \text{Stem}(X_{\text{origin}}) + \text{Stem}(X_{\text{origin}} + X_{\text{diff}}), \label{eq1} \end{equation} where $X_{\text{diff}}$ represents differences between consecutive frames.

A combination of 2D-CNNs, batch normalization, ReLU activation, and max-pooling is used to enhance these features. Finally, a 5×5 convolution and self-attention mechanism adaptively focus on informative facial regions, improving feature quality for subsequent modules.

\subsection{Multi-Scale Synergistic State Space Duality (SSSD)}

The SSSD module combines State Space Models (SSMs) with attention mechanisms to capture both long-term dependencies and short-term variations in temporal data between the two pathways, as illustrated in Fig. \ref{PhysMamba frame}(a). A key component of SSSD is the Multi-Scale Query (MQ) mechanism, which allows the model to process information across multiple temporal scales, as shown in Fig. \ref{PhysMamba frame}(b). By ensuring efficient information exchange between the pathways, MQ enhances the model's adaptability to diverse and dynamic input characteristics, such as motion artifacts and inconsistent lighting. This design allows PhysMamba to excel in both time-domain and frequency-domain tasks, ensuring high performance in real-world scenarios.

To simplify, SSMs operate like "memory systems" that track evolving patterns over time. They are mathematically described as: \begin{align} h_{t} = A h_{t-1} + B x_{t}, \quad y_{t} = C^{\top} h_{t}, \label{eq_ssm} \end{align} where $A$, $B$, and $C$ are learnable matrices controlling state transitions and feature outputs.

Building on this, the State Space Duality (SSD) framework introduces structured attention mechanisms to efficiently encode temporal dependencies: \begin{equation} \mathrm{SSD} = (L \circ Q K^{\top}) \cdot V, \label{eq_ssd} \end{equation} 
Through the MQ mechanism, the query ($Q$), key ($K$), and value ($V$) matrices are mapped from inputs of different temporal scales, with $L$ as a structured mask matrix that replaces traditional softmax operations.

The SSSD extends this concept by enabling complementary interactions between pathways:

The SA Pathway focuses on self-dependencies within individual features: \begin{equation} X_{\text{SA}} = (L_{S} \circ Q_{S} K_{S}^{\top}) \cdot V_{S}. \label{eq_sa} \end{equation}
The CA pathway receives Multi-Scale $Q_{S}$ from SA, enabling CA to promote cross-feature refinement.: \begin{equation} X_{\text{CA}} = (L_{C} \circ Q_{S} K_{C}^{\top}) \cdot V_{C}. \label{eq_ca} \end{equation}
As illustrated in Fig. \ref{PhysMamba frame}(c), these pathways function like two analysts: one focuses on deeply examining individual data streams (SA), while the other identifies meaningful interactions across them (CA). This collaboration ensures robust temporal feature learning, even under noisy conditions.

\subsection{Frequency Domain Feed-Forward (FDF)}

To enhance periodic physiological patterns, the FDF network applies Fast Fourier Transform (FFT) to convert temporal signals $h(t)$ into the frequency domain: \begin{align} H(f) = \int_{-\infty}^{+\infty} h(t) e^{-j2\pi f t} dt. \label{eq_fft} \end{align} This emphasizes periodic components like heart rate signals. The inverse FFT restores the frequency-enhanced signals to the temporal domain, preserving these enhancements. Outputs from the SA and CA pathways, denoted as $F_{SA}$ and $F_{CA}$, are combined to optimize rPPG signal representation.

\begin{table*}[t]
\caption{Intra-dataset results on UBFC, PURE, and MMPD datasets.}
\label{tab:intra}
\centering
\large
\resizebox{\textwidth}{!}{
\begin{tabular}{lcccc|cccc|ccccc}
\hline
\textbf{Method} & \multicolumn{4}{c|}{\textbf{UBFC}} & \multicolumn{4}{c|}{\textbf{PURE}} & \multicolumn{5}{c}{\textbf{MMPD}} \\
\cline{2-14}
& MAE & RMSE & MAPE & $r$ & MAE & RMSE & MAPE & $r$ & MAE & RMSE & MAPE & $r$ & SNR \\
\hline
GREEN \cite{verkruysse2008remote} & 19.73 & 31.00 & 18.72 & 0.37 & 10.09 & 23.85 & 10.28 & 0.34 & 21.68 & 27.69 & 24.39 & -0.01 & -14.34 \\
ICA \cite{poh2010non}           & 16.00 & 25.65 & 15.35 & 0.44 & 4.77  & 16.07 & 4.47  & 0.72 & 18.60 & 24.30 & 20.88 & 0.01  & -13.84 \\
CMROM \cite{de2013robust}          & 4.06  & 8.83  & 3.34  & 0.89 & 5.77  & 14.93 & 11.52 & 0.81 & 13.66 & 18.76 & 16.00 & 0.08  & -11.74 \\
LGI  \cite{pilz2018local}          & 15.80 & 28.55 & 14.70 & 0.36 & 4.61  & 15.38 & 4.96  & 0.77 & 17.08 & 23.32 & 18.98 & 0.04  & -13.15 \\
PBV  \cite{de2014improved}          & 15.90 & 26.40 & 15.17 & 0.48 & 3.92  & 12.99 & 4.84  & 0.84 & 17.95 & 23.58 & 20.18 & 0.09  & -13.88 \\
POS   \cite{wang2016algorithmic}         & 4.08  & 7.72  & 3.93  & 0.92 & 3.67  & 11.82 & 7.25  & 0.88 & 12.36 & 17.71 & 14.43 & 0.18  & -11.53 \\
OMIT   \cite{casado2023face2ppg}        & 15.79 & 28.54 & 14.69 & 0.36 & 4.65  & 15.81 & 4.96  & 0.75 & 7.80  & 12.00 & 10.55 & 0.13  & -8.68  \\
\hline
DeepPhys  \cite{chen2018deepphys}     & 0.76  & 1.09  & 0.79  & 0.99 & 3.33  & 14.45 & 2.91  & 0.90 & 23.73 & 28.25 & 25.63 & -0.06 & -15.45 \\
PhysNet  \cite{yu2019remote}      & 0.58  & 0.83  & 0.61  & 0.99 & 0.54  & 0.93  & 0.58  & 0.99 & 4.81  & 11.83 & 4.84  & 0.60  & 1.51   \\
TS-CAN  \cite{liu2020multi}       & 0.81  & 1.10  & 0.84  & 0.99 & 0.40  & 0.73  & 0.44  & 0.99 & 8.97  & 16.58 & 9.43  & 0.44  & -6.92  \\
PhysFormer  \cite{yu2022physformer}   & 0.63  & 0.98  & 0.65  & 0.99 & 0.25  & 0.37  & 0.34  & 0.99 & 13.64 & 19.39 & 14.42 & 0.15  & -11.02 \\
EfficientPhys \cite{liu2023efficientphys} & 0.72  & 1.01  & 0.75  & 0.99 & 5.10  & 16.61 & 4.19  & 0.87 & 12.79 & 21.12 & 13.48 & 0.24  & -9.23  \\
RhythmMamba \cite{zou2024rhythmmamba}& 0.54 & 0.79 & 0.54 & 0.99 & 0.29  & 0.39  & 0.36  & 0.99 & 3.16  & 7.27  & 3.37  & 0.84  & 4.74   \\
\textbf{PhysMamba (Ours)} & \textbf{0.45} & \textbf{0.76} & \textbf{0.45} & \textbf{0.99} & \textbf{0.24} & \textbf{0.36} & \textbf{0.31} & \textbf{0.99} & \textbf{2.84} & \textbf{6.41} & \textbf{3.04} & \textbf{0.88} & \textbf{5.20} \\
\hline
\end{tabular}
}
\label{intra}
\end{table*}

\subsection{rPPG Predictor}

The outputs from the SA and CA pathways are concatenated along the channel dimension: \begin{equation} X_{\text{fusion}} = \text{Concat}(F_{CA}, F_{SA}). \label{eq_fusion} \end{equation} A 1D convolutional layer then refines the fused features to produce the final rPPG predictions. By integrating both time-domain and frequency-domain features, the rPPG Predictor ensures accurate heart rate estimation even under challenging conditions, such as varying lighting and motion artifacts.

\section{Experiments}

\subsection{Datasets}

To rigorously evaluate PhysMamba's performance, we selected three widely-used rPPG datasets with varying levels of complexity: UBFC-rPPG \cite{bobbia2019unsupervised}, PURE \cite{stricker2014non}, and MMPD \cite{tang2023mmpd}. These datasets present distinct challenges, ranging from controlled laboratory conditions to highly complex real-world scenarios involving significant noise and motion artifacts.

\textbf{UBFC-rPPG}: This dataset consists of videos from 42 subjects performing mathematical tasks to induce heart rate variability. The relatively clean facial videos with minimal noise make it a benchmark for assessing model accuracy under simple conditions.

\textbf{PURE}: Comprising videos of 10 subjects under six controlled head movement scenarios—including stillness, talking, and slow/fast rotations—PURE introduces moderate motion artifacts, testing the model's robustness to dynamic conditions.

\textbf{MMPD}: A highly challenging dataset featuring 33 subjects under varying lighting conditions, skin tones, and head movement complexities. Capturing over 11 hours of video with diverse environmental factors, MMPD includes significant noise and motion artifacts, simulating real-world complexities. Due to computational constraints, we utilized the mini-MMPD version for experimentation.

We evaluated PhysMamba using standard metrics: Mean Absolute Error (MAE), Root Mean Squared Error (RMSE), Mean Absolute Percentage Error (MAPE), and Pearson correlation coefficient ($r$). For MMPD, we also computed the Signal-to-Noise Ratio (SNR) to assess robustness in high-noise environments.

\begin{figure}[t] \centering \includegraphics[width=0.9\linewidth]{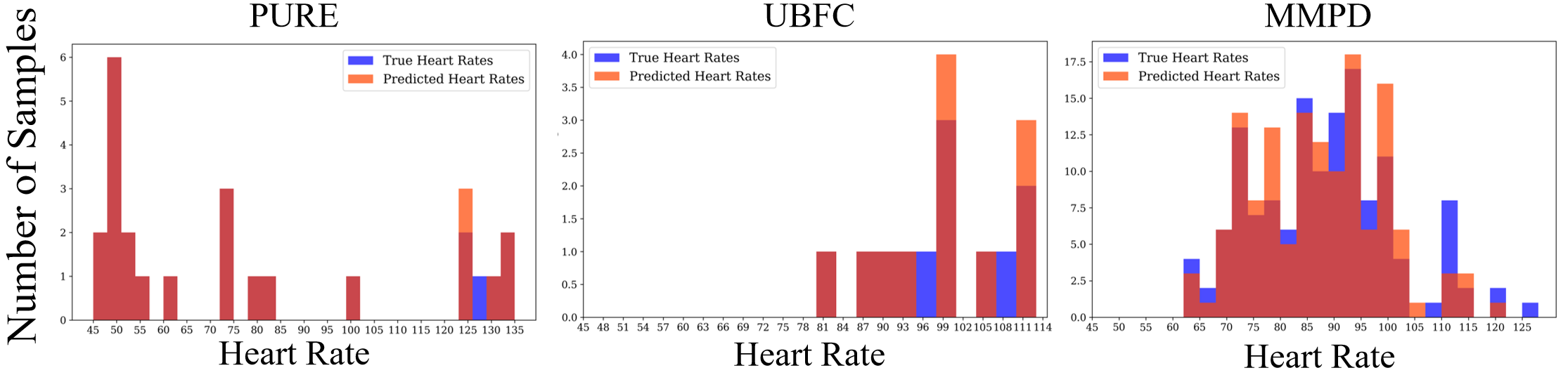} \caption{Comparison of predicted and ground-truth heart rate distributions across UBFC-rPPG, PURE, and MMPD datasets.} \label{fig:HR} \end{figure}

\subsection{Experimental Details}

All experiments were conducted using the open-source \textit{rppg-toolbox} \cite{liu2024rppg} in PyTorch. Videos were preprocessed by cropping and resizing facial regions to $128 \times 128$ pixels. The model was trained using a learning rate of $3 \times 10^{-4}$, a batch size of 16, and for 30 epochs. The SSD model's dimension and state space size were set to 64, with a head dimension of 16. All computations were performed on an NVIDIA RTX 4090 GPU.

The loss function combined time-domain loss ($L_{\text{Time}}$) and frequency-domain loss ($L_{\text{Freq}}$) for robust learning:

\begin{equation}
\mathcal{L}_{\text{overall}} = L_{\text{Time}} + L_{\text{Freq}},
\end{equation}

where $L_{\text{Time}}$ minimizes the negative Pearson correlation between the predicted and ground-truth rPPG signals, and $L_{\text{Freq}}$ aligns their frequency-domain representations.

\begin{figure}[t]
\centering
\includegraphics[scale=0.17]{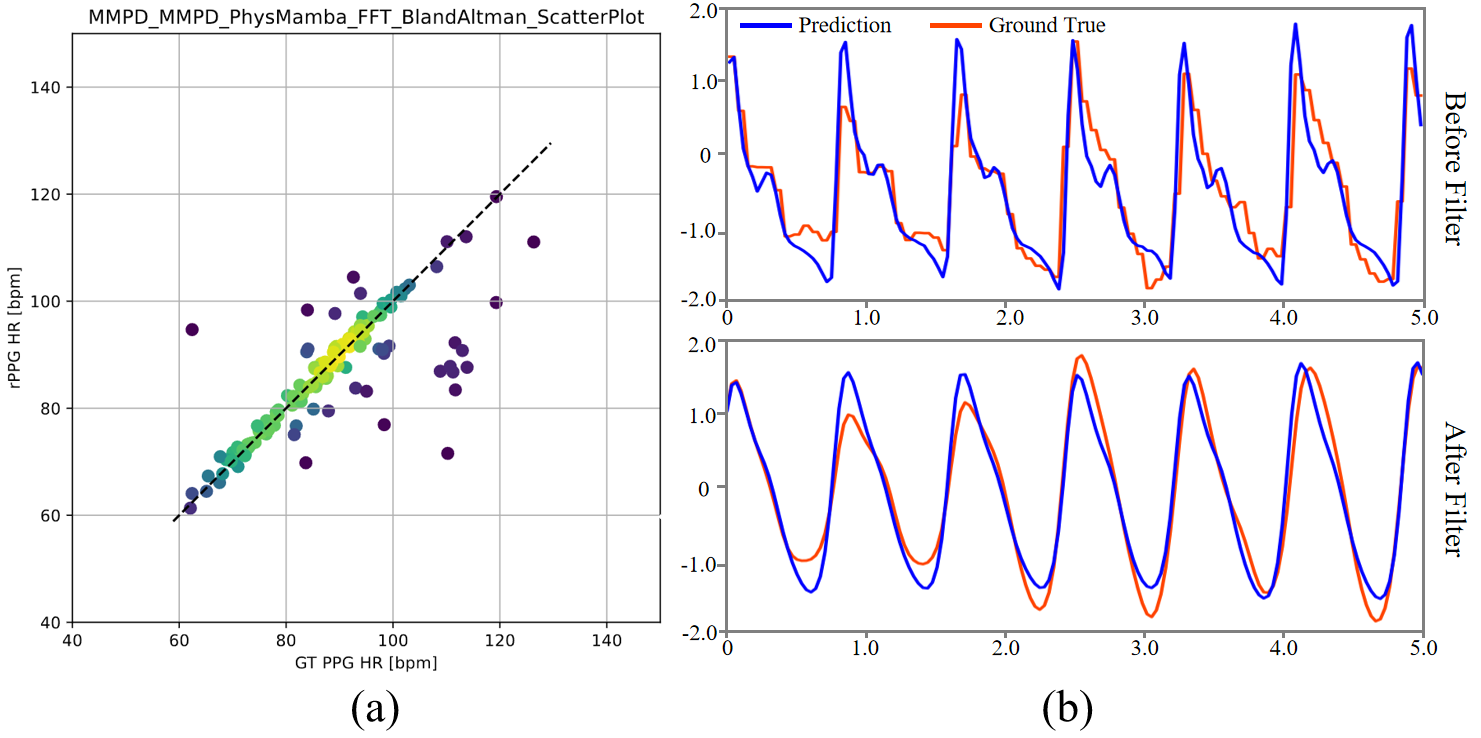}
\caption{Visualization of the results from the MMPD dataset. (a) the Bland-Altman plot. (b) the waveform of a sample before and after filtering.}
\label{visual}
\end{figure}

\section{Experimental Evaluation}

To fully verify PhysMamba's effectiveness and superiority in rPPG heart rate estimation, we have designed a detailed evaluation process, including both intra-dataset and cross-dataset assessments. We compared PhysMamba with Unsupervised methods including GREEN\cite{verkruysse2008remote}, ICA\cite{poh2010non}, CHROM\cite{de2013robust}, LGI\cite{pilz2018local}, PBV\cite{de2014improved}, and POS\cite{wang2016algorithmic}, OMIT\cite{casado2023face2ppg}, as well as state-of-the-art deep learning methods including DeepPhys\cite{chen2018deepphys}, PhysNet\cite{yu2019remote}, TS-CAN\cite{liu2020multi}, PhysFormer\cite{yu2022physformer},EfficientPhys\cite{liu2023efficientphys}, and RhythmMamba\cite{zou2024rhythmmamba}. We provide additional experimental details in the supplementary materials.

\subsection{Intra-Dataset Testing}

In intra-dataset evaluations, PhysMamba was trained and tested on the same dataset to assess its performance under consistent conditions. As shown in Table \ref{tab:intra}, PhysMamba achieved near-optimal results on UBFC-rPPG and PURE datasets, with MAEs of 0.45 bpm and 0.24 bpm, respectively, outperforming existing methods. On the challenging MMPD dataset, PhysMamba demonstrated superior robustness, achieving an MAE of 2.84 bpm and the highest SNR of 5.20. These results indicate PhysMamba's ability to handle noise, motion artifacts, and varying lighting conditions effectively.The low computational complexity of PhysMamba, enabled by the SSSD framework, makes it particularly suitable for deployment on mobile devices and real-time monitoring systems. Its robustness on MMPD highlights its capability to adapt to complex, noisy environments.

\subsection{Cross-Dataset Testing}

We selected the PURE dataset with head movements and the MMPD dataset with significant noise. Cross-dataset evaluations assessed PhysMamba's generalization by training and testing on different datasets, a crucial factor for real-world applications where training and deployment data often differ.

As shown in Tables \ref{tab:MMPD_PURE} and \ref{tab:PURE_MMPD}, PhysMamba consistently outperformed existing methods in cross-dataset scenarios. When trained on MMPD and tested on PURE, it achieved an MAE of 5.32 bpm, surpassing RhythmMamba. Conversely, when trained on PURE and tested on MMPD, it achieved an MAE of 9.87 bpm, significantly better than other methods.

The superior generalization of PhysMamba can be attributed to: \textbf{SSSD} : By combining state space models with attention mechanisms between the two pathways, PhysMamba effectively captures both long-term dependencies and short-term variations. This dual capability enhances the model's robustness against dataset shifts caused by lighting, skin tone, and motion variability.
\textbf{Multi-Scale Queries}: MQ enables efficient feature sharing across different time scales, improving the model's ability to adapt to diverse data distributions. This is particularly evident in the cross-dataset scenarios, where PhysMamba maintains high accuracy despite significant domain shifts.
\textbf{Practical Implications:} The ability to generalize across datasets demonstrates PhysMamba's potential for deployment in real-world environments with minimal retraining. This is especially valuable for applications like remote health monitoring, where training data may not cover all possible variations in deployment settings.

\begin{figure} \centering \includegraphics[width=0.9\linewidth]{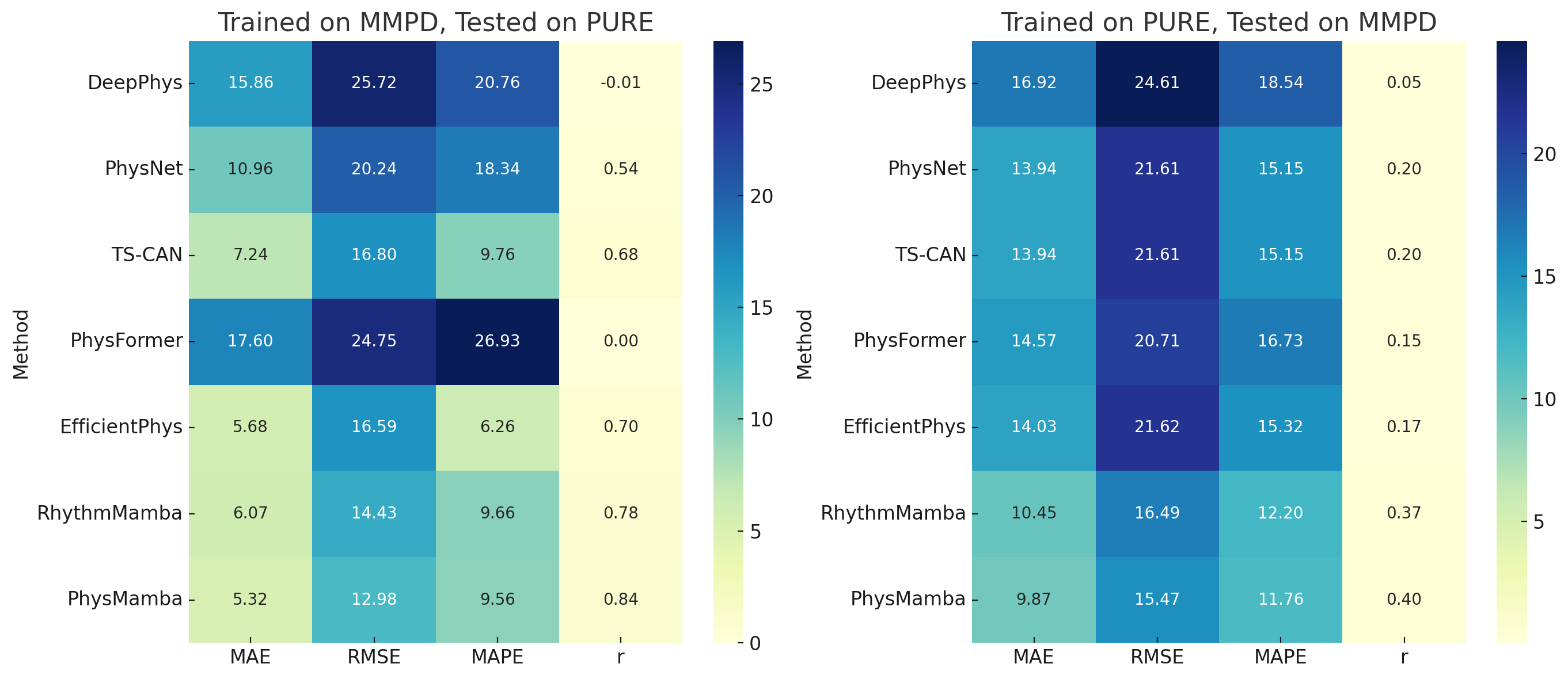} \caption{Heat map of cross-dataset results: Training on MMPD and testing on PURE (left), and vice versa (right). PhysMamba shows superior generalization compared to other methods.} \label{fig:cross} \end{figure}

\begin{table}[t]
\caption{Cross-dataset results: Trained on MMPD, tested on PURE.}
\centering
\fontsize{9pt}{11pt}\selectfont
\setlength{\tabcolsep}{4pt}
\begin{tabular}{lcccc}
\hline
\textbf{Method} & \textbf{MAE} & \textbf{RMSE} & \textbf{MAPE} & $r$ \\
\hline
DeepPhys \cite{chen2018deepphys} & 15.86 & 25.72 & 20.76 & -0.01 \\
PhysNet \cite{yu2019remote} & 10.96 & 20.24 & 18.34 & 0.54 \\
TS-CAN \cite{liu2020multi} & 7.24 & 16.80 & 9.76 & 0.68 \\
PhysFormer \cite{yu2022physformer} & 17.60 & 24.75 & 26.93 & 0.00 \\
EfficientPhys \cite{liu2023efficientphys} & 5.68 & 16.59 & \textbf{6.26} & 0.70 \\
RhythmMamba \cite{zou2024rhythmmamba}& 6.07 & 14.43 & 9.66 & 0.78  \\
\textbf{PhysMamba (Ours)} & \textbf{5.32} & \textbf{12.98} & 9.56 & \textbf{0.84} \\
\hline
\end{tabular}
\label{tab:MMPD_PURE}
\end{table}
\begin{table}[t]
\caption{Cross-dataset results: Trained on PURE, tested on MMPD.}
\centering
\fontsize{9pt}{11pt}\selectfont
\setlength{\tabcolsep}{4pt}
\begin{tabular}{lcccc}
\hline
\textbf{Method} & \textbf{MAE} & \textbf{RMSE} & \textbf{MAPE} & $r$ \\
\hline
DeepPhys \cite{chen2018deepphys} & 16.92 & 24.61 & 18.54 & 0.05 \\
PhysNet \cite{yu2019remote} & 13.94 & 21.61 & 15.15 & 0.20 \\
TS-CAN \cite{liu2020multi} & 13.94 & 21.61 & 15.15 & 0.20 \\
PhysFormer \cite{yu2022physformer} & 14.57 & 20.71 & 16.73 & 0.15 \\
EfficientPhys \cite{liu2023efficientphys} & 14.03 & 21.62 & 15.32 & 0.17 \\
RhythmMamba \cite{zou2024rhythmmamba}& 10.45 & 16.49 & 12.20 & 0.37 \\
\textbf{PhysMamba (Ours)} & \textbf{9.87} & \textbf{15.47} & \textbf{11.76} & \textbf{0.40} \\
\hline
\end{tabular}
\label{tab:PURE_MMPD}
\end{table}

\begin{table}[t]
\caption{Ablation study results on the MMPD dataset. MQ: Multi-Scale Queries, FDF: Frequency Domain Feed-Forward, DP: Dual-Pathway, SSD: State Space Duality, CA: Cross-Attention.}
\centering
\fontsize{9pt}{11pt}\selectfont
\setlength{\tabcolsep}{3pt}
\begin{tabular}{cccccccccc}
\hline
\textbf{MQ} & \textbf{FDF} & \textbf{DP} & \textbf{SSD} & \textbf{CA} & \textbf{MAE} & \textbf{RMSE} & \textbf{MAPE} & $r$ & \textbf{SNR} \\
\hline
\checkmark & \checkmark &  &  &  & 4.40 & 8.81 & 4.66 & 0.76 & 3.36 \\
\checkmark & \checkmark &  & \checkmark &  & 3.46 & 7.21 & 3.69 & 0.84 & 3.94 \\
\checkmark & \checkmark & \checkmark &  &  & 3.13 & 6.55 & 3.31 & 0.87 & 4.63 \\
\checkmark & \checkmark & \checkmark & \checkmark &  & 3.72 & 8.06 & 3.95 & 0.80 & 4.15 \\
\checkmark &  & \checkmark & \checkmark & \checkmark & 3.53 & 7.46 & 3.83 & 0.82 & 1.50 \\
 & \checkmark & \checkmark & \checkmark & \checkmark & 3.49 & 7.35 & 3.69 & 0.83 & 4.17 \\
\hline
\checkmark & \checkmark & \checkmark & \checkmark & \checkmark & \textbf{2.84} & \textbf{6.41} & \textbf{3.04} & \textbf{0.88} & \textbf{5.20} \\
\hline
\end{tabular}
\label{Ablation_Study}
\end{table}

\subsection{Ablation Study}

To evaluate the contribution of each key component, we conducted ablation studies on the MMPD dataset. The results, summarized in Table \ref{Ablation_Study}, highlight the importance of the MQ, FDF, SSD, and CA modules in achieving state-of-the-art performance.

Multi-Scale Queries (MQ): Improved the model's ability to capture multi-scale dependencies, reducing the MAE from 4.40 to 3.46 bpm.
Frequency Domain Feed-Forward (FDF): Enhanced the detection of periodic patterns, increasing SNR from 1.50 to 5.20.
Dual-Pathway Design (DP): Reduced redundancy and improved feature learning, achieving an MAE of 3.13 bpm.
SSSD Framework: Enabled robust temporal learning, ensuring the best overall performance with an MAE of 2.84 bpm and an SNR of 5.20.
\textbf{Practical Implications:} The ablation study underscores the complementary roles of PhysMamba's components, particularly MQ and FDF, in balancing computational efficiency with robustness, making the model ideal for resource-constrained environments.

\subsection{Visualization and Further Analysis}

To further analyze PhysMamba's performance, Figure \ref{fig:HR} compares the predicted and ground-truth heart rate distributions across the three datasets. The results show a close alignment, even in challenging conditions like MMPD.
In Figure \ref{visual}, we present the visualization of the MMPD results, showing a strong correlation between the predicted and the ground truth heart rate.A second-order Butterworth filter (cutoff frequencies: 0.75 and 2.5 Hz) filtered the predicted PPG waveform. This result confirms PhysMamba's accuracy and reliability in heart rate estimation, even in challenging scenarios, demonstrating strong robustness.
Figure \ref{fig:cross} highlights the cross-dataset results, showing PhysMamba's consistent performance across different training and testing conditions. This further validates its robustness and generalization capabilities. These visualizations reinforce the potential of PhysMamba for real-world applications.

% These visualizations reinforce the potential of PhysMamba for real-world applications. Its ability to maintain high accuracy across datasets highlights its robustness in handling unseen conditions, such as varying lighting, skin tones, and motion artifacts.

\section{Conclusion}
\label{conclusion}
\vspace{0.04in}
This paper introduces PhysMamba, an efficient rPPG estimation method using State Space Duality in a dual-pathway time-frequency network. It combines long-term dependency modeling with a full-sequence attention mechanism. We also propose a simple yet innovative Synergistic State Space Duality algorithm with Multi-Scale Queries, facilitating effective information interaction within the dual-pathway network. Both internal and cross-dataset tests show that PhysMamba achieves state-of-the-art performance, with high accuracy and robustness, indicating strong potential in real-world physiological and psychological applications. 

Future work will focus on extending PhysMamba for real-time monitoring of psychological states, emotions, and overall health in dynamic and diverse environments. With its high efficiency and adaptability, PhysMamba provides a scalable foundation for advancing non-contact health monitoring technologies, benefiting both individual well-being and public healthcare systems.

% \bibliographystyle{IEEEbib}
% \bibliography{icme2025references}
% Generated by IEEEtran.bst, version: 1.14 (2015/08/26)

\vspace{12pt}
% \color{red}
% IEEE conference templates contain guidance text for composing and formatting conference papers. Please ensure that all template text is removed from your conference paper prior to submission to the conference. Failure to remove the template text from your paper may result in your paper not being published.

\end{document}